\documentclass[11pt,twocolumn]{article}
\usepackage[utf8]{inputenc}
\usepackage[T1]{fontenc}
\usepackage{amsmath,amsfonts,amssymb}
\usepackage{graphicx}
\usepackage{booktabs}
\usepackage{url}
\usepackage{natbib}
\usepackage{authblk}
\usepackage[margin=1in]{geometry}
\usepackage{xcolor}
\usepackage{algorithm}
\usepackage{algorithmic}

\title{\Large \textbf{Characterizing Fitness Landscape Structures in Prompt Engineering}}

\author[1,2]{Arend Hintze} 
\affil[1]{Department of Data Analytics, Dalarna University, Sweden,  \texttt{ahz@du.se},
\texttt{ORCID: 0000-0002-4872-1961}}
\affil[2]{BEACON Institute for the Study of Evolution in Action, Michigan State University, USA}

\date{\today}

\begin{document}

\maketitle

\begin{abstract}
While prompt engineering has emerged as a crucial technique for optimizing large language model performance, the underlying optimization landscape remains poorly understood. Current approaches treat prompt optimization as a black-box problem, applying sophisticated search algorithms without characterizing the landscape topology they navigate. We present a systematic analysis of fitness landscape structures in prompt engineering using autocorrelation analysis across semantic embedding spaces. Through experiments on error detection tasks with two distinct prompt generation strategies -- systematic enumeration (1,024 prompts) and novelty-driven diversification (1,000 prompts) -- we reveal fundamentally different landscape topologies. Systematic prompt generation yields smoothly decaying autocorrelation, while diversified generation exhibits non-monotonic patterns with peak correlation at intermediate semantic distances, indicating rugged, hierarchically structured landscapes. Task-specific analysis across 10 error detection categories reveals varying degrees of ruggedness across different error types. Our findings provide an empirical foundation for understanding the complexity of optimization in prompt engineering landscapes.
\end{abstract}

\section{Introduction}

Prompt engineering has emerged as the primary interface between human intent and large language model capabilities, fundamentally shaping how we extract desired behaviors from these powerful systems \citep{brown2020language, liu2023pre}. The practice spans from casual users iteratively refining ChatGPT queries to researchers systematically optimizing prompts for complex reasoning tasks \citep{wei2022chain, suzgun2022challenging}. Remarkably, much prompt engineering practice—both human-driven iteration and automated optimization—operates under an implicit assumption about the optimization landscape: that small modifications to prompts yield proportionally small changes in performance. This assumption underlies the commonly reported practice of iterative prompt refinement -- making incremental adjustments to wording, structure, or examples in the expectation of gradual improvement \citep{zamfirescu2023johnny}. In computer science, this same assumption forms the theoretical foundation for hill-climbing algorithms, gradient-based optimization, and local search methods \citep{russell2020artificial}.

This intuitive assumption has driven the evolution of prompt engineering from manual trial-and-error approaches to sophisticated automated optimization methods. Current automated approaches broadly fall into three categories: gradient-based methods that leverage model internals to compute discrete "gradients" through techniques like AutoPrompt's token substitution \citep{shin2020autoprompt} and continuous prompt tuning \citep{li2021prefix, lester2021power}; discrete search methods that treat models as black boxes, including beam search approaches \citep{pryzant2023automatic} and large language models as prompt engineers \citep{zhou2022large}; and evolutionary approaches that balance exploration and exploitation through population-based search \citep{guo2024connecting, fernando2023promptbreeder}. Yet despite the critical importance of the smooth landscape assumption for both human intuition and algorithmic design, the actual topology of prompt optimization landscapes remains entirely uncharacterized.

Early empirical evidence challenges the smooth landscape assumption. Lu et al. \citep{lu2022fantastically} demonstrated that simply reordering a few-shot example can cause 30+ point performance variations in specific tasks, while Zhao et al. \citep{zhao2021calibrate} showed extreme sensitivity to seemingly minor wording choices. These findings suggest that at least some prompt engineering contexts may exhibit rugged landscape characteristics -- with many local optima, fitness cliffs, and non-linear relationships between prompt similarity and performance -- rather than uniformly smooth topologies. This possibility has begun to emerge in both academic literature and industry practice, with practitioners frequently reporting optimization challenges that suggest complex landscape topology: promising prompt modifications that lead to performance dead-ends, semantically similar prompts with vastly different capabilities, and the surprising effectiveness of population-based search methods over gradient-like approaches.

Despite the prevalence of prompt engineering practice, systematic empirical research on how humans actually navigate prompt spaces remains surprisingly limited. While automated optimization methods receive extensive investigation, there are no systematic studies of human iterative refinement patterns, semantic distance preferences, or longitudinal optimization strategies. Manual prompt engineering is often reported as labor-intensive and challenging, yet the field lacks a fundamental understanding of the behavioral patterns that may drive human optimization decisions. This gap between the prevalence of reported practice and the paucity of systematic investigation creates both a theoretical puzzle and a practical opportunity for understanding the optimization challenges that practitioners may face.

The theoretical framework for understanding such complex optimization domains is rooted in fitness landscape analysis, which originated in evolutionary biology with Sewall Wright's adaptive landscape metaphor \citep{wright1932roles} and has become central to understanding the difficulty of optimization problems in evolutionary computation. The NK model \citep{kauffman1989nk} provides a tunable framework for generating landscapes with varying degrees of ruggedness through epistatic interactions between problem variables, while autocorrelation analysis \citep{weinberger1990correlated} quantifies landscape structure by measuring how performance correlation decays with distance, enabling classification of landscapes from smooth to rugged. Recent applications of landscape analysis have provided valuable insights into optimization complexity across machine learning contexts, including neural architecture search \citep{white2021powerful}, hyperparameter optimization \citep{schneider2022hpo}, and deep learning loss surfaces \citep{li2018visualizing, garipov2018loss}. However, these applications focus on continuous parameter spaces or discrete architectural choices, leaving the unique characteristics of semantic text optimization entirely unexplored. Drawing on this established framework as a potentially useful lens, we investigate whether fitness landscape analysis can provide insights into prompt engineering optimization complexity.

The convergence of these two insights—that practitioners may operate under assumptions about optimizable landscapes while empirical evidence suggests complex, rugged topologies in some contexts—creates both a theoretical puzzle and a practical opportunity. If prompt engineering indeed violates the smooth landscape assumption underlying incremental refinement approaches, then the theoretical framework developed by Kauffman, Weinberger, and others may provide precisely the analytical tools needed to characterize this complexity. The reported success of evolutionary algorithms in recent prompt optimization work \citep{guo2024connecting, fernando2023promptbreeder} provides suggestive indirect evidence supporting this hypothesis: population-based search methods typically outperform local optimization precisely when landscapes exhibit the rugged, epistatic structure that fitness landscape analysis was designed to quantify. However, this inference from algorithmic performance alone leaves the actual landscape topology uncharacterized, treating the optimization space as an opaque domain rather than a structured mathematical object amenable to systematic analysis.

Addressing these questions requires moving beyond black-box optimization to direct landscape characterization. This endeavor faces a fundamental methodological challenge that has constrained the broader field: the absence of objective evaluation criteria for most natural language tasks. While comprehensive surveys have catalogued the rapid proliferation of prompt engineering techniques—with over 58 distinct methods identified for large language models alone \citep{schulhoff2024prompt}—the evaluation landscape remains fragmented and largely subjective. As noted in systematic reviews of the field, existing prompt engineering methods typically require significant amounts of labeled data, access to model parameters, or both \citep{sorensen2022information}, while automated metrics developed for earlier NLP tasks often fail to capture the nuances of modern generative outputs.

The evaluation challenge is particularly acute for prompt engineering research because it operates at the intersection of two inherently subjective domains: natural language generation quality and task-specific performance assessment. Unlike mathematics, programming, or formal logic—where correctness can be definitively established—most language tasks exist in a continuous space of quality gradations. Creative writing, summarization, dialogue generation, and open-ended reasoning lack the clear binary success criteria that enable systematic optimization analysis. Recent efforts to establish unified evaluation frameworks \citep{zhu2023promptbench} have made progress in standardizing assessment protocols, but the fundamental challenge remains: how to conduct rigorous landscape analysis when the fitness function itself resists precise quantification.

Our approach addresses this evaluation challenge through a concrete framework using error detection tasks that provide clear binary outcomes. We created pairs of statements where one version is correct and another contains a specific error. For example, a correct statement "The Great Wall of China was built primarily during the Ming Dynasty" becomes incorrect as "The Great Wall of China was built primarily during the Tang Dynasty" (factual error), or "The Great Wall of China were built primarily during the Ming Dynasty" (grammatical error). This design enables systematic evaluation: an LLM can correctly identify a true statement (true positive), correctly flag an actual error (true positive), incorrectly report a failure where none exists (false positive), or overlook a real error (false negative). By testing across 10 error categories—logical reasoning, mathematical computation, syntactic analysis, factual verification, and others—we create a diverse yet objectively evaluable landscape that enables systematic optimization analysis while maintaining the binary correctness criteria essential for quantitative landscape characterization.

Through systematic fitness landscape reconstruction, we address the fundamental question: are prompt optimization landscapes smooth and amenable to incremental refinement, or rugged and resistant to local search? Our analysis reveals a striking finding that challenges simple characterizations of prompt engineering complexity. For error detection tasks, the landscape topology is not an intrinsic property of the task domain, but rather depends critically on how prompts are generated and what regions of semantic space are explored. Systematic categorical enumeration produces relatively smooth, navigable landscapes that support incremental optimization intuitions, while semantic diversification reveals complex, hierarchically structured topologies with non-monotonic correlation patterns that violate traditional optimization assumptions. This method-dependent landscape structure provides the first empirical foundation for understanding why different optimization approaches succeed or fail in this domain, with implications for both theoretical understanding and practical optimization strategy design in prompt engineering.

\section{Methodology}

\subsection{Error Detection Task Framework}

We constructed a dataset of 100 test cases using ChatGPT-4o-mini, with 10 test cases for each of 10 error categories: (1) Lexical/Word Choice (inappropriate vocabulary), (2) Grammatical/Syntactic (agreement, tense, structure), (3) Logic (inconsistencies, contradictions), (4) Orthographic/Spelling (misspellings, capitalization), (5) Content Omission (missing information), (6) Content Addition (irrelevant information), (7) Mathematical (calculation errors, impossible numbers), (8) Programming (syntax, algorithmic errors), (9) Physics (violations of physical laws), and (10) Factual/Knowledge (incorrect claims). Each test case includes a correct statement and an identical statement with one error introduced, with duplicate statements removed to ensure unique evaluation instances.

\subsection{Dual-LLM Evaluation Framework}

We implemented a dual-LLM system using Llama 3.2 models to ensure objective evaluation. LLM1 (Generator) receives the test prompt and statement, attempting error detection with temperature 0.3. LLM2 (Evaluator) compares the generator's response against ground truth using statement alignment rather than re-solving the original task, operating at temperature 0.1 for consistency.

The evaluator assigns scores on a three-point scale: +1 (response aligns well with ground truth), 0 (partially correct but incomplete), -1 (contradicts ground truth). These scores map to the true/false positive/negative framework described earlier. For partial correctness (score 0), we distribute 0.25 to each confusion matrix category, acknowledging the uncertainty inherent in ambiguous responses while avoiding arbitrary binary classifications.

\subsection{Prompt Space Exploration}

Given the limited empirical understanding of how humans navigate prompt optimization spaces, we employ two distinct exploration strategies designed to systematically characterize landscape topology through different sampling approaches. This dual-strategy methodology enables us to distinguish between sampling-dependent artifacts and fundamental landscape properties while providing comprehensive coverage of the prompt semantic space.

The systematic approach ensures complete categorical coverage by systematically perturbing all possible emphasis combinations across error categories, while the novelty-driven approach employs algorithmic diversity maximization to achieve relatively homogeneous distribution across the broader semantic space. By comparing landscape characteristics across these fundamentally different sampling strategies, we can identify which topological features represent intrinsic properties of the optimization landscape versus artifacts of particular exploration patterns.

\subsubsection{Systematic Prompt Generation}

To ensure comprehensive categorical coverage, we generated all possible combinations of the 10 error categories ($2^{10} = 1,024$ prompts) using the template: ``You are a helpful assistant, check the next statement, pay attention to [category combinations]''. This systematic enumeration approach systematically perturbs all possible emphasis combinations across error categories, providing complete coverage of the categorical prompt space while maintaining semantic coherence. This strategy enables analysis of landscape structure within a constrained but thoroughly explored semantic region.

\subsubsection{Novelty-Driven Diversification}

To achieve relatively homogeneous distribution across the broader semantic space while avoiding clustering artifacts, we employed an advanced novelty search algorithm adapted for semantic text optimization \citep{lehman2011abandoning, mouret2015illuminating}. This approach generated 1,000 diverse prompts optimized for embedding space coverage rather than categorical completeness, using algorithmic diversity maximization to ensure well-distributed sampling across semantic regions. Novelty search, originally developed by Lehman and Stanley, abandons traditional fitness-based optimization in favor of rewarding behavioral diversity, making it particularly well-suited for exploring semantic spaces where traditional gradient-based approaches may get trapped in local optima or produce heterogeneous clustering.

The algorithm employs:
\begin{itemize}
\item \textbf{Embedding-based novelty scoring} using all-MiniLM-L6-v2
\item \textbf{Reservoir-based optimization} with k=10 nearest neighbors  
\item \textbf{Constrained generation} maintaining semantic focus on error detection
\end{itemize}

\begin{algorithm}
\caption{Novelty-Driven Prompt Generation}
\begin{algorithmic}[1]
\STATE Initialize reservoir $R$ with seed prompt
\STATE Set reservoir size limit $N_{max} = 256$
\FOR{round = 1 to 1000}
    \STATE $p_{parent} \leftarrow$ random selection from $R$
    \STATE $p_{new} \leftarrow$ generate\_variation($p_{parent}$)
    \STATE $emb_{new} \leftarrow$ embed($p_{new}$)
    \STATE $neighbors \leftarrow$ k-nearest neighbors of $emb_{new}$ in $R$
    \STATE $novelty \leftarrow$ mean distance to $neighbors$
    \IF{$|R| < N_{max}$}
        \STATE Add $p_{new}$ to $R$
    \ELSIF{$novelty >$ min\_novelty in $R$}
        \STATE Replace least novel prompt in $R$ with $p_{new}$
    \ENDIF
\ENDFOR
\RETURN $R$
\end{algorithmic}
\end{algorithm}

This approach systematically explores diverse regions of the semantic space while maintaining task relevance, generating prompts that span a wide range of linguistic formulations for error detection instructions.

\subsection{Autocorrelation Analysis}

We measure landscape ruggedness through autocorrelation analysis, a fundamental technique from evolutionary computation that quantifies how performance correlation varies with semantic distance in embedding space. Autocorrelation analysis addresses a core question in optimization: if two solutions are similar, how likely are their performance values to be similar as well?

In a \textbf{smooth landscape}, nearby solutions tend to have similar performance—small changes in prompts lead to small, predictable changes in effectiveness. For example, in a smooth landscape we would expect changing "check this statement" to "verify this statement" to yield similar error detection performance. Autocorrelation in smooth landscapes shows high correlation at short distances that gradually decays as distance increases, reflecting the intuitive notion that "similar inputs produce similar outputs."

In a \textbf{rugged landscape}, this relationship breaks down. Semantically similar prompts can have dramatically different performance—in a rugged landscape we might observe that changing "analyze" to "examine" unexpectedly halves the error detection rate, while switching to a completely different approach like "find problems in" paradoxically improves performance. Rugged landscapes show rapid, erratic correlation decay with little predictable relationship between prompt similarity and performance similarity.

The autocorrelation function $\rho(d)$ measures the Pearson correlation between fitness values of prompt pairs separated by semantic distance $d$. For prompt engineering, this reveals whether the optimization landscape respects our intuitions about semantic similarity: Do prompts that "mean similar things" actually perform similarly? Or does the landscape exhibit the kind of complex, epistatic structure that makes incremental optimization unreliable?

Understanding landscape topology is crucial because it determines which optimization strategies will succeed. Smooth landscapes favor local search and gradient-based methods, while rugged landscapes require population-based approaches like evolutionary algorithms. By characterizing prompt landscape structure for error detection tasks, we can predict a priori which optimization approaches are most likely to find effective prompts for given tasks in this domain.

\subsection{Simulated Manual Prompt Engineering}

We implemented distance-constrained random walk optimization to measure achievable performance improvements under varying semantic step size limitations. The experimental protocol identified the 50 worst-performing prompts from each generation strategy based on overall accuracy scores as optimization starting points. From each starting prompt, we performed 100 independent random walks where each step moves to a randomly selected neighbor within a specified semantic distance threshold.

We established a unified semantic distance framework covering the complete range observed across both prompt generation strategies: systematic (0.002 to 0.385) and novelty-driven (0.102 to 1.225) cosine distance units. Testing proceeded across 50 equally-spaced thresholds spanning the combined range (0.002 to 1.225).

Each random walk follows this procedure: starting from a worst-performing prompt, identify all prompts within the distance threshold of the current position, randomly select a neighbor from available options, and move to that location regardless of fitness change. We track the maximum fitness encountered during each walk. The algorithm terminates after 50 steps, when no improvement occurs for 10 consecutive steps, or when no neighbors exist within the distance threshold.

For each distance threshold, we computed the average maximum fitness achieved across all 5,000 optimization runs (50 starting points × 100 walks) and the standard deviation to assess outcome variability. This protocol generates optimization curves showing how step size limitations affect achievable performance improvements across different landscape topologies.

\section{Results}

\subsection{Performance Distribution Characteristics}

Our analysis begins with examining the performance distributions produced by each prompt generation strategy, revealing fundamentally different exploration patterns across the optimization landscape.

\begin{figure*}[t]
\centering
\includegraphics[width=5.5in]{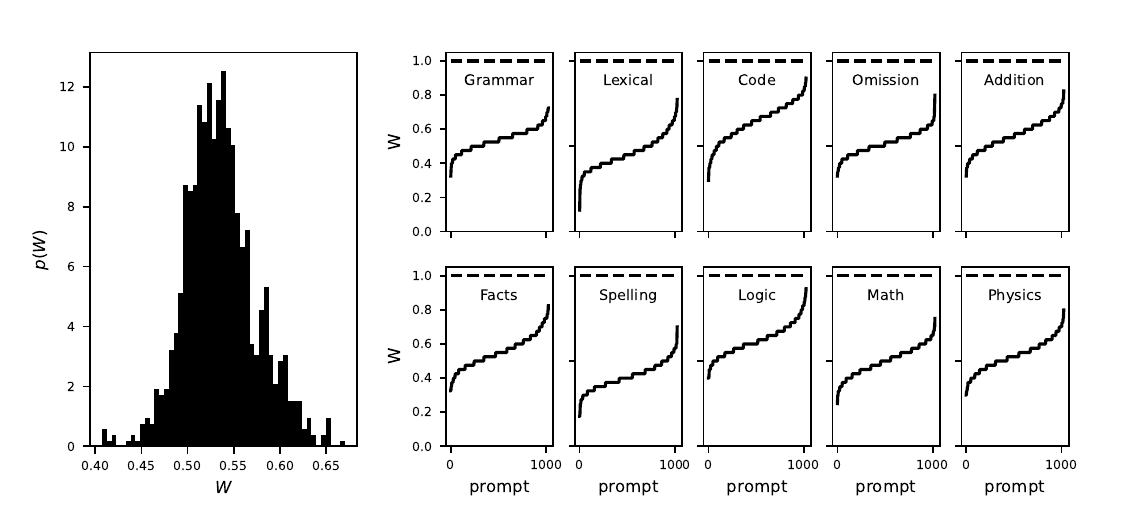}
\caption{Systematic categorical enumeration produces narrow performance distributions. The 1,024 prompts generated through systematic combination of error categories show concentrated performance clustered around 0.5 accuracy with limited variance.}
\label{fig:systematic}
\end{figure*}

The systematic approach (Figure \ref{fig:systematic}) generates a narrow, concentrated performance distribution tightly clustered around 0.5 accuracy. This concentrated distribution suggests that systematic categorical combinations, while comprehensive in their coverage of error category permutations, converge toward a consistent performance band with limited variance.

\begin{figure*}[t]
\centering
\includegraphics[width=5.5in]{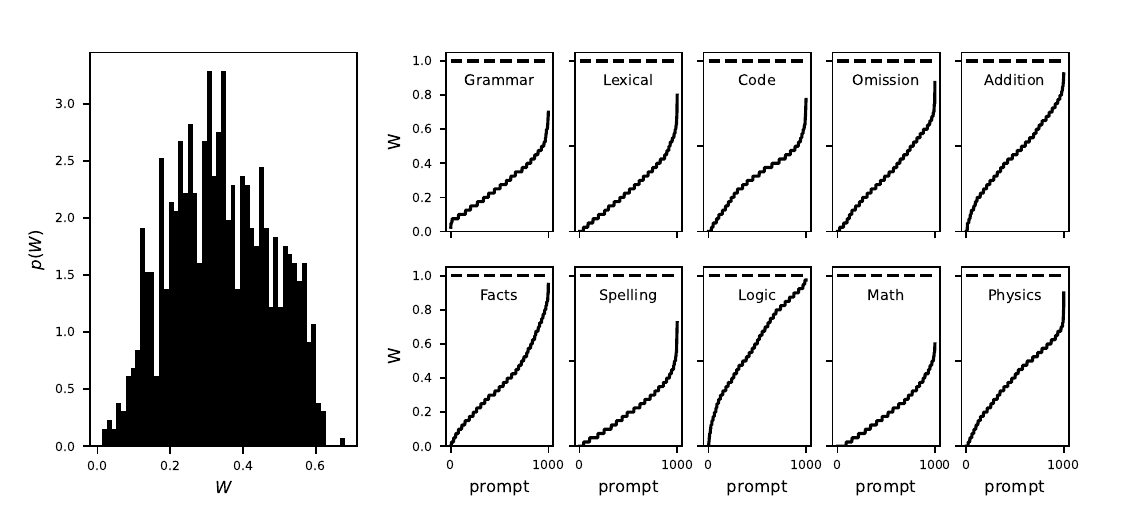}
\caption{Novelty-driven diversification generates broad, multimodal performance distributions. The 1,000 prompts produced through semantic diversification show wide performance variance spanning the entire range from 0.0 to 0.6+ accuracy with clear multimodal characteristics.}
\label{fig:novelty}
\end{figure*}

In contrast, the novelty-driven approach (Figure \ref{fig:novelty}) produces a broad performance distribution spanning the entire performance range from near-zero to over 0.6 accuracy, with clear multimodal characteristics. This wide variance indicates that novelty search explores diverse areas of the fitness landscape, accessing both very low-performing regions and potentially higher-performance peaks that systematic enumeration fails to reach.

\subsection{Landscape Visualization and Autocorrelation Analysis}

\begin{figure}[t]
\centering
\includegraphics[width=\columnwidth]{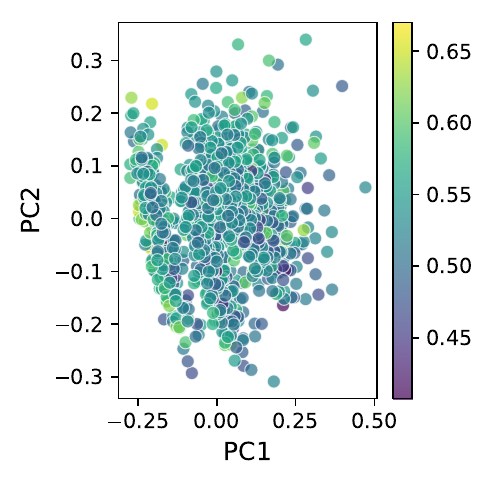}
\caption{PCA visualization of systematic categorical enumeration prompts. Each point represents a prompt in the first two principal components, with colors indicating performance values. The visualization fails to reveal meaningful landscape structure.}
\label{fig:pca-systematic}
\end{figure}

\begin{figure}[t]
\centering
\includegraphics[width=\columnwidth]{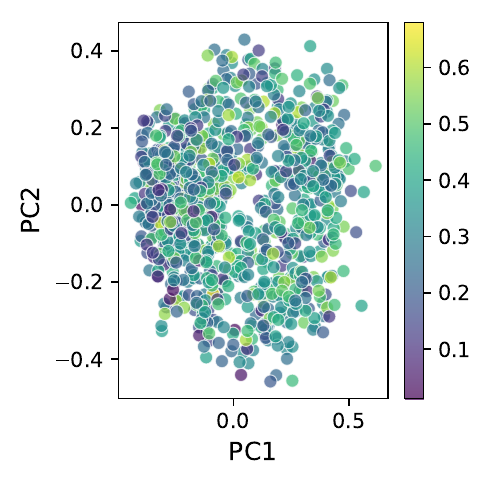}
\caption{PCA visualization of novelty-driven diversification prompts. Points represent prompts projected onto the first two principal components, colored by performance. The visualization provides little insight into landscape topology.}
\label{fig:pca-novelty}
\end{figure}

Conventional landscape visualization through Principal Component Analysis provides limited insights into the fitness landscape structure. Figures \ref{fig:pca-systematic} and \ref{fig:pca-novelty} show PCA projections of prompt embeddings with performance values encoded as colors, but neither visualization provides clear actionable insights into the relationship between prompt similarity and performance correlation.

\begin{figure}[t]
\centering
\includegraphics[width=\columnwidth]{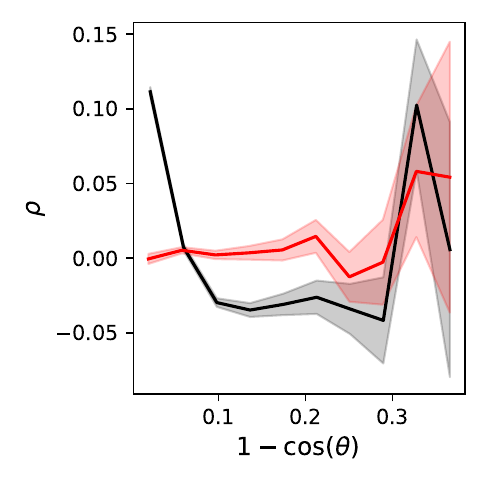}
\caption{Autocorrelation analysis for systematic categorical enumeration (1,024 prompts). The systematic approach shows smooth landscape topology with predictable correlation decay.}
\label{fig:autocorr-systematic}
\end{figure}

\begin{figure}[t]
\centering
\includegraphics[width=\columnwidth]{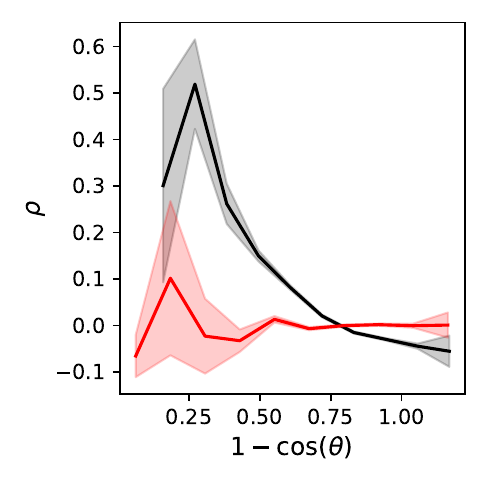}
\caption{Autocorrelation analysis for novelty-driven diversification (1,000 prompts). The novelty approach reveals rugged landscape structure with non-monotonic correlation pattern showing peak correlation at intermediate semantic distances.}
\label{fig:autocorr-novelty}
\end{figure}

Autocorrelation analysis reveals the true landscape structure that PCA visualization cannot capture. For systematic categorical enumeration (Figure \ref{fig:autocorr-systematic}), we observe a smooth landscape with predictable decay structure. The novelty-driven approach (Figure \ref{fig:autocorr-novelty}) exhibits a rugged landscape with a striking non-monotonic autocorrelation pattern: low correlation at short distances, peak correlation at intermediate distances (~0.3), then decay at longer distances.

\subsection{Task-Specific Landscape Analysis}

\begin{figure}[t]
\centering
\includegraphics[width=2in]{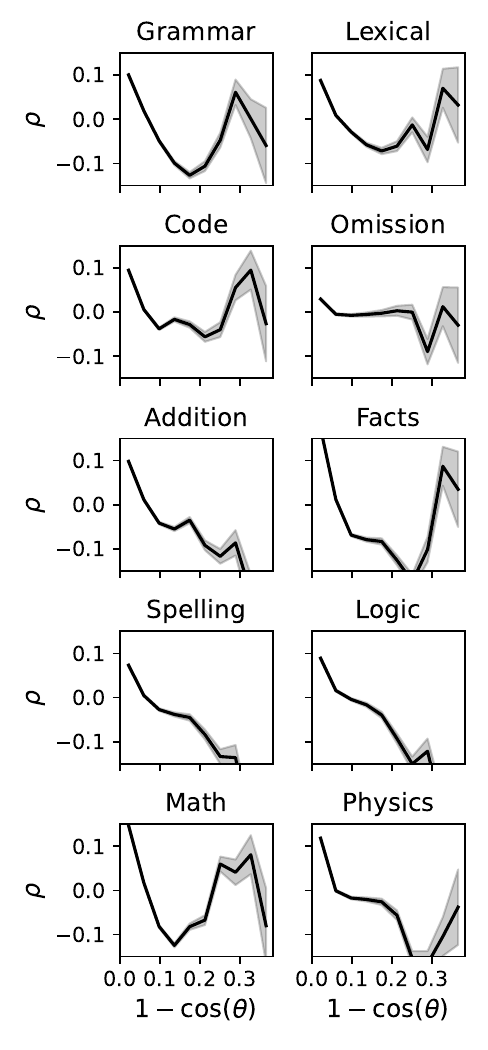}
\caption{Task-specific autocorrelation analysis for systematic categorical enumeration across 10 error detection categories. Different error types show distinct landscape topologies: some categories exhibit smooth decay patterns consistent with navigable optimization landscapes, while others demonstrate non-monotonic patterns.}
\label{fig:autocorr-categories-systematic}
\end{figure}

Analysis across the 10 error detection categories reveals systematic variation in landscape ruggedness across different task types. The systematic approach (Figure \ref{fig:autocorr-categories-systematic}) shows variable landscape topologies across error categories, with some domains exhibiting smooth decay patterns that allow incremental "tinkering" optimization, while others demonstrate more complex structures suggesting that small changes do not yield predictable improvements.

\begin{figure}[t]
\centering
\includegraphics[width=2in]{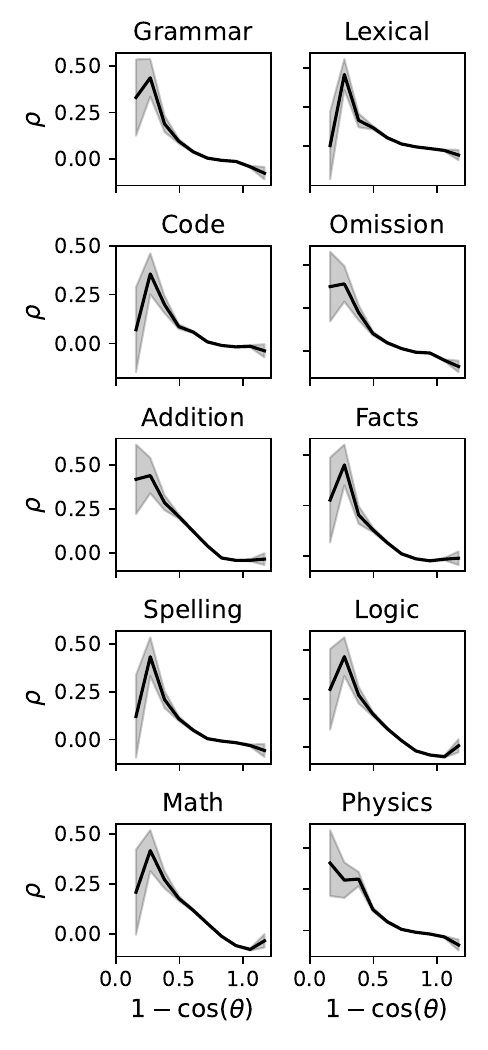}
\caption{Task-specific autocorrelation analysis for novelty-driven diversification across 10 error detection categories. Nearly all categories show consistent non-monotonic patterns with peak correlation at intermediate distances (~0.3), indicating universal ruggedness across task types.}
\label{fig:autocorr-categories-novelty}
\end{figure}

The novelty-driven approach (Figure \ref{fig:autocorr-categories-novelty}) reveals remarkably consistent non-monotonic patterns across most categories. Nearly every error detection category exhibits the characteristic correlation "hump" with peak correlation at intermediate distances (~0.3), with the possible exception of content omission. This consistency suggests that the rugged landscape structure with optimal correlation at intermediate semantic distances is a consistent pattern across different error types when the full prompt space is explored through semantic diversification.

\subsection{Optimization Difficulty Validation}

Autocorrelation analysis characterizes the static structure of fitness landscapes by measuring performance correlation as a function of semantic distance. However, the relationship between static landscape structure and dynamic optimization behavior requires separate investigation. When practitioners engage in prompt "tinkering"—iteratively refining prompts through successive modifications—the semantic magnitude of these changes constrains which regions of the landscape become accessible. The systematic and novelty-driven generation strategies produce prompts spanning different semantic distance ranges, suggesting that optimization under realistic step size constraints may behave differently across these landscapes.

To investigate how semantic step size limitations affect optimization effectiveness, we modeled prompt improvement as distance-constrained random walks starting from poor-performing prompts. This approach simulates optimization scenarios where practitioners make bounded semantic changes, testing whether the structural differences revealed by autocorrelation analysis translate to measurable optimization difficulty differences. The experiment examines how maximum achievable performance varies with step size constraints across both landscape types.

\begin{figure}[t]
\centering
\includegraphics[width=\columnwidth]{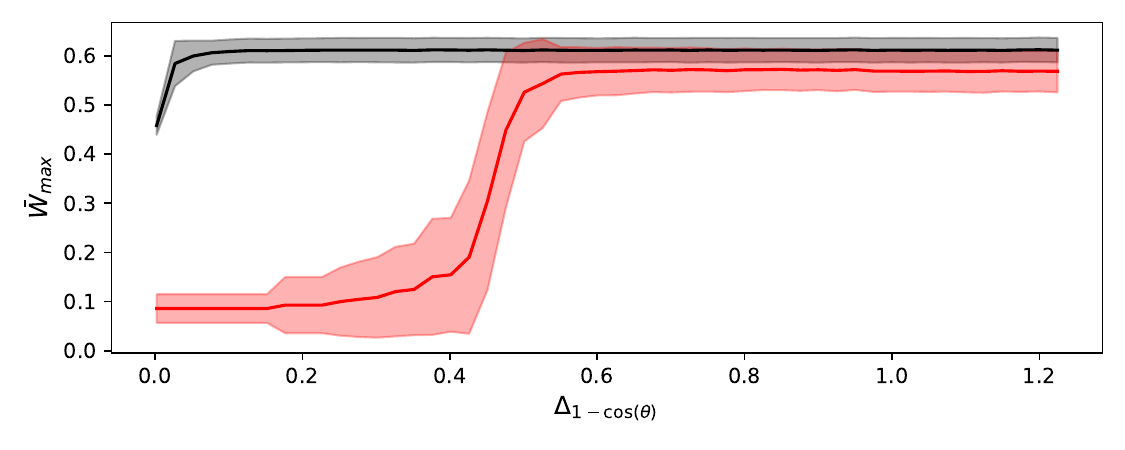}
\caption{Optimization difficulty comparison between systematic categorical enumeration and novelty-driven diversification landscapes. Each curve shows the average maximum fitness achieved across 5,000 optimization runs (50 starting points × 100 random walks) as a function of maximum semantic step size. Error bars indicate standard deviation. The systematic landscape (black) shows rapid improvement with small steps followed by plateau behavior, while the novelty landscape (red) exhibits poor performance at small step sizes with dramatic improvement. Standard deviation across the samples is given as shadows begind the means.}
\label{fig:optimization-difficulty}
\end{figure}

Figure \ref{fig:optimization-difficulty} presents the experimental results, revealing dramatically different optimization behaviors that precisely confirm the autocorrelation predictions. The systematic landscape demonstrates classic smooth optimization characteristics: rapid improvement from 0.458 to 0.61 accuracy occurs with small semantic steps (distance threshold $\leq$ 0.063), followed by plateau behavior across larger step sizes. This validates the effectiveness of incremental refinement approaches in systematically generated prompt spaces, where semantic similarity reliably predicts performance similarity.

The novelty-driven landscape exhibits fundamentally different optimization pathology that validates the rugged landscape hypothesis. Performance remains trapped at low levels (~0.086 accuracy) across small step sizes, demonstrating that minor semantic modifications cannot escape local optima. The dramatic improvement beginning around distance threshold 0.4-0.5, with performance jumping to ~0.54 accuracy, occurs precisely in the intermediate distance range where autocorrelation analysis revealed peak correlation. This empirical correspondence between predicted and observed optimal step sizes provides compelling validation that fitness landscape analysis can predict effective optimization strategies for prompt engineering.

The optimization curves establish a direct causal link between landscape topology and practical optimization difficulty in this experimental context. The systematic landscape's smooth structure enables successful hill-climbing with small semantic changes, while the novelty landscape's rugged topology requires larger semantic jumps to access regions where incremental improvement becomes possible. This finding transforms autocorrelation analysis from a descriptive tool into a predictive framework for prompt optimization strategy selection in error detection tasks.

The different semantic distance ranges accessed by each generation strategy (systematic: 0.002-0.385, novelty: 0.102-1.225) indicate that sampling method determines which regions of prompt space are explored. The systematic approach explores a constrained subset of the semantic landscape, while novelty-driven generation accesses broader semantic diversity. This suggests that optimization method effectiveness may depend on the specific prompt space region being explored, with implications for understanding why different optimization approaches succeed in different contexts.

\section{Discussion}

This study addressed a fundamental question about the nature of prompt engineering optimization: Are prompt landscapes smooth and amenable to incremental "tinkering" approaches, or do they exhibit rugged, complex topologies that require sophisticated search strategies? Our investigation reveals that the answer depends critically on how the prompt space is explored, with profound implications for both theoretical understanding and practical optimization approaches.

Our findings demonstrate that exploration strategy fundamentally shapes the observed landscape topology in error detection tasks using Llama 3.2. Systematic categorical enumeration produces relatively smooth landscapes with predictable autocorrelation decay, supporting the intuitive notion that semantically similar prompts yield similar performance within this constrained space. However, when the full semantic space is explored through novelty-driven diversification, a dramatically different picture emerges: rugged landscapes characterized by a striking non-monotonic autocorrelation pattern with peak correlation at intermediate semantic distances (~0.3 cosine distance).

This non-monotonic organization challenges the basic assumptions underlying current prompt engineering practice. Rather than the expected pattern where nearby prompts perform similarly and distant prompts show no correlation, we observe a complex structure: low correlation at short distances, peak correlation at intermediate distances, and decorrelation at long distances. This pattern appears consistently across nearly all error detection categories when semantic diversification is employed, suggesting a consistent property of prompt optimization landscapes in this domain rather than task-specific phenomena.

Critically, traditional landscape visualization methods fail entirely to reveal this structure. Principal Component Analysis, despite being the standard approach for high-dimensional optimization visualization, provides no actionable insights into the relationship between prompt similarity and performance correlation. Only through autocorrelation analysis—borrowed from evolutionary computation theory—does the true landscape organization become apparent. This methodological insight has implications beyond prompt engineering, suggesting that fitness landscape analysis techniques may be broadly applicable to understanding optimization challenges in semantic spaces.

The discovery that optimal predictability occurs at intermediate rather than short semantic distances fundamentally contradicts the theoretical foundation of incremental optimization approaches. This finding suggests a possible explanation for why human "polishing" of prompts sometimes fails to yield expected improvements and why population-based search methods may outperform gradient-like approaches in prompt optimization. Small changes in prompt wording appear to create unpredictable noise rather than systematic improvement, while systematically different but semantically related approaches provide more reliable optimization pathways.

\section{Limitations and Future Work}

This study presents several important limitations that define the scope of our findings and highlight directions for future research.

\subsection{Model Architecture and Scale Dependencies}

Our analysis focuses specifically on Llama 3.2 models, and the observed landscape characteristics may not generalize across different model architectures or scales. Larger language models or alternative architectures may exhibit fundamentally different landscape topologies due to varying training procedures, emergent capabilities, and internal representations. The relationship between model scale and landscape structure represents a critical area for future investigation, as optimization strategies that prove effective for smaller models may fail entirely when applied to frontier-scale systems.

\subsection{Training Data and Foundation Model Dependencies}

The landscape structures we observe are inherently tied to the specific training procedures and datasets used for Llama 3.2. Different training regimes, fine-tuning approaches, or foundation model families may produce entirely different optimization landscapes. This dependency suggests that landscape analysis may need to be conducted separately for each model family, limiting the generalizability of optimization insights across the rapidly evolving language model ecosystem.

\subsection{Task Domain Constraints}

While error detection provides clear evaluation criteria that enable systematic landscape analysis, generalization to other natural language tasks remains an open question. Creative generation, open-ended reasoning, dialogue systems, and domain-specific applications may exhibit entirely different landscape characteristics that cannot be inferred from binary classification tasks. The trade-off between evaluation objectivity and task representativeness represents a fundamental challenge for systematic prompt optimization research.

\subsection{Limited Understanding of Human Optimization Behavior}

Our theoretical framework assumes that human practitioners engage in incremental refinement based on semantic similarity intuitions, but systematic empirical research on actual human prompt optimization patterns remains scarce. We lack comprehensive studies of how humans navigate prompt spaces, what semantic distances they typically explore, and whether their optimization strategies align with the landscape structures we identify. The behavioral assumptions underlying our work—that humans engage in "tinkering" behavior and expect incremental improvements from small changes—are based on anecdotal reports rather than systematic empirical investigation. This gap between computational landscape analysis and empirically validated human optimization strategies represents a significant limitation for translating our findings into practical guidance.

\subsection{Future Research Directions}

Several critical research directions emerge from these limitations:

\textbf{Cross-Model Landscape Analysis}: Systematic comparison of landscape structures across different model architectures, scales, and training approaches to understand the generalizability of optimization principles.

\textbf{Human Behavior Studies}: Longitudinal empirical research tracking how human practitioners actually navigate prompt spaces, including analysis of semantic distances, iteration patterns, and success strategies.

\textbf{Task Diversity Extension}: Development of systematic landscape analysis methods for subjective and open-ended tasks, potentially through human preference modeling or multi-objective optimization frameworks.

\textbf{Temporal Dynamics}: Investigation of how landscape structures evolve with model updates, fine-tuning, and changing capabilities, including the stability of optimization strategies over time.

\textbf{Optimization Strategy Development}: Translation of landscape insights into practical optimization algorithms that account for the hierarchical structure and non-monotonic correlation patterns we identify.

\textbf{Benchmark Development}: Creation of standardized evaluation frameworks for prompt optimization research that balance objective assessment with task diversity and practical relevance.

\section{Conclusion}

This work presents the first systematic characterization of fitness landscape structures in prompt engineering for error detection tasks, revealing complex, method-dependent topologies that challenge simple assumptions about optimization complexity. Using Llama 3.2 models, our findings demonstrate that landscape structure is not an intrinsic property of the task domain, but rather depends critically on the prompt generation strategy employed. Systematic categorical enumeration produces relatively smooth landscapes amenable to incremental optimization, while semantic diversification reveals rugged, hierarchically organized topologies with non-monotonic autocorrelation patterns.

These results have immediate implications for prompt optimization research within the error detection domain and suggest broader questions for the field. By moving beyond black-box optimization to direct landscape characterization, we enable more principled approaches to understanding optimization complexity in semantic spaces. The hierarchical organization we identify—with distinct micro-, meso-, and macro-level structures—provides both theoretical insight into semantic optimization properties and empirical evidence that peak correlation occurs at intermediate rather than short semantic distances.

Our work establishes fitness landscape analysis as a valuable framework for understanding prompt engineering complexity in evaluable domains. While our findings are specific to error detection tasks using Llama 3.2, they suggest that the theoretical tools developed for evolutionary optimization may provide precisely the analytical foundation needed to characterize optimization challenges in language model interfaces. However, significant questions remain about how these computational insights translate to actual human optimization strategies, how findings generalize across different model architectures and task domains, and whether similar landscape structures emerge in other areas of natural language interaction.

The discovery that optimal predictability occurs at intermediate semantic distances offers a counterintuitive but empirically grounded principle for systematic prompt exploration in this domain. Future work bridging computational landscape analysis with empirical studies of human optimization behavior will be essential for translating these theoretical insights into practical guidance for the broader prompt engineering community.



\section*{COI}
The author declares no conflict of interest
\section*{Data Availability}
Code and Prompt data is available at: \url{https://osf.io/trusw/}
\section*{Generative AI}
Generative AI (Claude, Sonnet 4, and Opus 4) was used to edit and improve text and code; however, the author approves all writing and considers generative AI as a tool, not as something that has any authoring right or autonomy towards that end.
\bibliographystyle{plainnat}
\bibliography{references}






\end{document}